\documentclass{article}

\usepackage{microtype}
\usepackage{graphicx}
\usepackage{subfigure}
\usepackage{booktabs} 

\usepackage{hyperref}

\usepackage[preprint]{icml2026}

\usepackage{amsmath}
\usepackage{amssymb}
\usepackage{mathtools}
\usepackage{amsthm}

\usepackage[capitalize,noabbrev]{cleveref}

\theoremstyle{plain}

\theoremstyle{definition}

\theoremstyle{remark}

\usepackage[textsize=tiny]{todonotes}

\usepackage[utf8]{inputenc}
\usepackage[most]{tcolorbox} 
\usepackage{xcolor}          

\definecolor{mybluebg}{RGB}{225, 230, 255}    
\definecolor{myblueframe}{RGB}{180, 190, 255} 
\definecolor{myblueheader}{RGB}{160, 175, 255}

\newtcolorbox{mybox}[1]{
    colback=mybluebg,           
    colframe=myblueframe,       
    colbacktitle=myblueheader,  
    coltitle=white,             
    fonttitle=\bfseries\sffamily, 
    title=#1,                   
    arc=5pt,                    
    outer arc=5pt,              
    left=10pt,                  
    right=10pt,                 
    top=8pt,                    
    bottom=8pt,                 
    boxrule=0.8pt,              
    titlerule=0pt,              
    toptitle=3pt,               
    bottomtitle=3pt             
}

\icmltitlerunning{Research on World Models Is Not Merely Injecting World Knowledge into Specific Tasks}

\begin{document}

\twocolumn[
\icmltitle{Research on World Models Is Not Merely Injecting \\ World Knowledge into Specific Tasks}



\icmlsetsymbol{equal}{*}

\begin{icmlauthorlist}
\icmlauthor{Bohan Zeng}{equal,yyy}
\icmlauthor{Kaixin Zhu}{equal,yyy}
\icmlauthor{Daili Hua}{equal,yyy}
\icmlauthor{Bozhou Li}{equal,yyy}
\icmlauthor{Chengzhuo Tong}{equal,yyy}
\icmlauthor{Yuran Wang}{equal,yyy}
\icmlauthor{Xinyi Huang}{equal,yyy}
\icmlauthor{Yifan Dai}{equal,sch1}
\icmlauthor{Zixiang Zhang}{equal,yyy}
\icmlauthor{Yifan Yang}{equal,yyy}
\icmlauthor{Zhou Liu}{yyy}
\icmlauthor{Hao Liang}{yyy}
\icmlauthor{Xiaochen Ma}{sch2}
\icmlauthor{Ruichuan An}{yyy}
\icmlauthor{Tianyi Bai}{sch2}
\icmlauthor{Hongcheng Gao}{sch4}
\icmlauthor{Junbo Niu}{yyy}
\icmlauthor{Yang Shi}{yyy}
\icmlauthor{Xinlong Chen}{sch3}
\icmlauthor{Yue Ding}{sch3}
\icmlauthor{Minglei Shi}{sch4}
\icmlauthor{Kai Zeng}{yyy}
\icmlauthor{Yiwen Tang}{yyy}
\icmlauthor{Yuanxing Zhang}{comp}
\icmlauthor{Pengfei Wan}{comp}
\icmlauthor{Xintao Wang}{comp}
\icmlauthor{Wentao Zhang}{yyy}
\end{icmlauthorlist}

\icmlaffiliation{yyy}{Peking University}
\icmlaffiliation{comp}{Kling Team, Kuaishou Technology}
\icmlaffiliation{sch1}{Shanghai Jiao Tong University}
\icmlaffiliation{sch2}{HKUST}
\icmlaffiliation{sch4}{Tsinghua University}
\icmlaffiliation{sch3}{School of Artificial Intelligence, University of Chinese Academy of Sciences}

\icmlcorrespondingauthor{Bohan Zeng}{bhzeng25@stu.pku.edu.cn}
\icmlcorrespondingauthor{Wentao Zhang}{wentao.zhang@pku.edu.cn}

\icmlkeywords{Machine Learning, ICML}

\vskip 0.3in
]



\printAffiliationsAndNotice{\icmlEqualContribution} 

\begin{abstract}
World models have emerged as a critical frontier in AI research, aiming to enhance large models by infusing them with physical dynamics and world knowledge. The core objective is to enable agents to understand, predict, and interact with complex environments. However, current research landscape remains fragmented, with approaches predominantly focused on injecting world knowledge into isolated tasks, such as visual prediction, 3D estimation, or symbol grounding, rather than establishing a unified definition or framework. While these task-specific integrations yield performance gains, they often lack the systematic coherence required for holistic world understanding. In this paper, we analyze the limitations of such fragmented approaches and propose a unified design specification for world models. We suggest that a robust world model should not be a loose collection of capabilities but a normative framework that integrally incorporates interaction, perception, symbolic reasoning, and spatial representation. This work aims to provide a structured perspective to guide future research toward more general, robust, and principled models of the world.
\end{abstract}

\begin{figure}[t]
    \centering
    \centerline{\includegraphics[width=0.95\columnwidth]{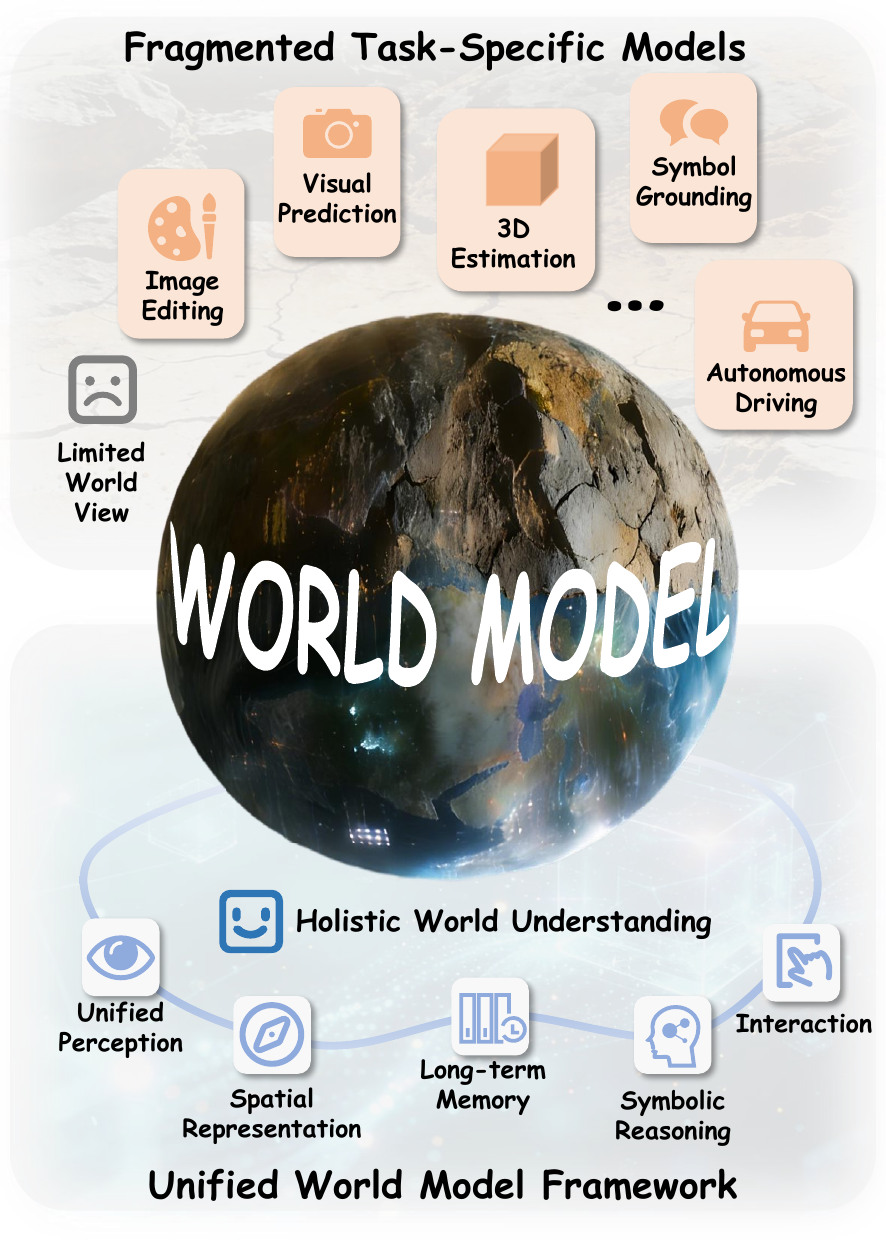}}
    \caption{Comparison between current task-specific paradigms and the proposed unified framework. While current research often reduces World Models to the injection of knowledge into specific tasks, a holistic World Model aims to endow AI with general capabilities to tackle multifaceted real-world challenges.}
    \label{fig:teaser}
    \vspace{-4mm}
\end{figure}

\section{Introduction}

With the explosive growth of internet data and continuous advances in neural network model training, existing large models~\cite{achiam2023gpt, bai2023qwen, yang2025qwen3, liu2024deepseek, bai2025qwen2, chen2024internvl, team2024gemini, lu2024deepseek} and diffusion models~\cite{liu2022flow, lipman2022flow, flux, peebles2023scalable} have achieved remarkable results in various fields. However, as model performance further improves, the bottleneck in data quality has become increasingly difficult to overcome, hindering further progress, especially in multimodal domains requiring precise analysis, such as multimodal reasoning, chemical formula recognition, 3D scene generation, and specific professional areas like healthcare~\cite{tang2024lgm, tang2025hunyuan, tochilkin2024triposr, xiang2025structured, xiang2025native, cheng2025mmaudio}. To break through the traditional token-prediction paradigm of large models, researchers have begun to focus on the study of world models.

The concept of World Model was first introduced by \cite{ha2018world}, which proposed a strategy of constructing an interactive system between agents and the world to handle complex visual input environments. With the rapid development of large models and various multimodal generation methods, recent work~\cite{zhu2024sora} has further expanded the notion of world models, viewing video generation and 3D generation as intelligent systems that simulate the real world. Researchers are considering world models as the next-generation paradigm to replace token-predicting large language models~\cite{yang2025cambrian}.

As world models attract growing interest, numerous research fields have begun to incorporate world knowledge to empower models to perform tasks that require an understanding of physical and contextual rules~\cite{yu2025wonderworld, team2025hunyuanworld, liu2025worldmirror, zhu2025astra}. This trend is evident in diverse applications, including image editing~\cite{zeng2025editworld, lin2025uniworld, chen2025unireal}, multimodal spatial reasoning~\cite{chen2024spatialvlm}, autonomous driving~\cite{tu2025role, zeng2025rethinking}, and even mobile communication methods such as MobileWorld~\cite{kong2025mobileworld}. Several studies ~\cite{hu2025simulating} have further provided systematic categorizations and summaries of these world-knowledge-infused generation methods, detailing their achieved capabilities.

However, these existing methods of injecting world knowledge into tasks still rely on fine-tuning models with human-curated, task-specific data. Even the most frontier and widely-discussed research at present remains this same pattern~\cite{openai2024sora, alibaba2025wan25, sun2025worldplay, russell2025gaia}. While this can improve performance on particular tasks, it does not break away from the inherent paradigm of the downstream tasks. Consequently, such approaches remain incapable of actively exploring, discovering, and responding to complex world environments, deviating from the original research objective of world models. The fundamental goal of a world model is to enable large models and agents to enhance their understanding of the complex world through active interaction with it, thereby making more accurate analyses and responses. Overemphasis on aligning the outputs of specific tasks with world rules may impede the development of world models.

To address these challenges and steer research toward a more holistic understanding of the physical world, this paper advocates for a shift from task-specific adaptations to a comprehensive system design. Specifically, the main contents and contributions of this work are organized as follows:

\begin{itemize}
    \item We provide a detailed review of recent progress in World Models, categorizing existing approaches into reasoning, content generation, and interactive agents. We examine how these fields currently incorporate world knowledge to enhance performance.
    
    \item We critically analyze the shortcomings of current methods that rely on injecting knowledge into isolated tasks. Through case studies in LLMs, video generation, and embodied AI, we demonstrate that these approaches often fail to achieve genuine physical understanding and long-term consistency.
    
    \item We propose a unified and standardized World Model Framework. We define the essential components, including Interaction, Reasoning, Memory, Environment, and Multimodal Generation, and articulate how they should be integrally designed to support robust world simulation.
    
    \item We identify critical directions for future breakthroughs, such as physically-grounded spatiotemporal representation, embodied interaction control, and autonomous modular evolution, to guide the community toward more general and principled models.
\end{itemize}

\begin{figure*}[h]
    \centering
    \centerline{\includegraphics[width=\linewidth]{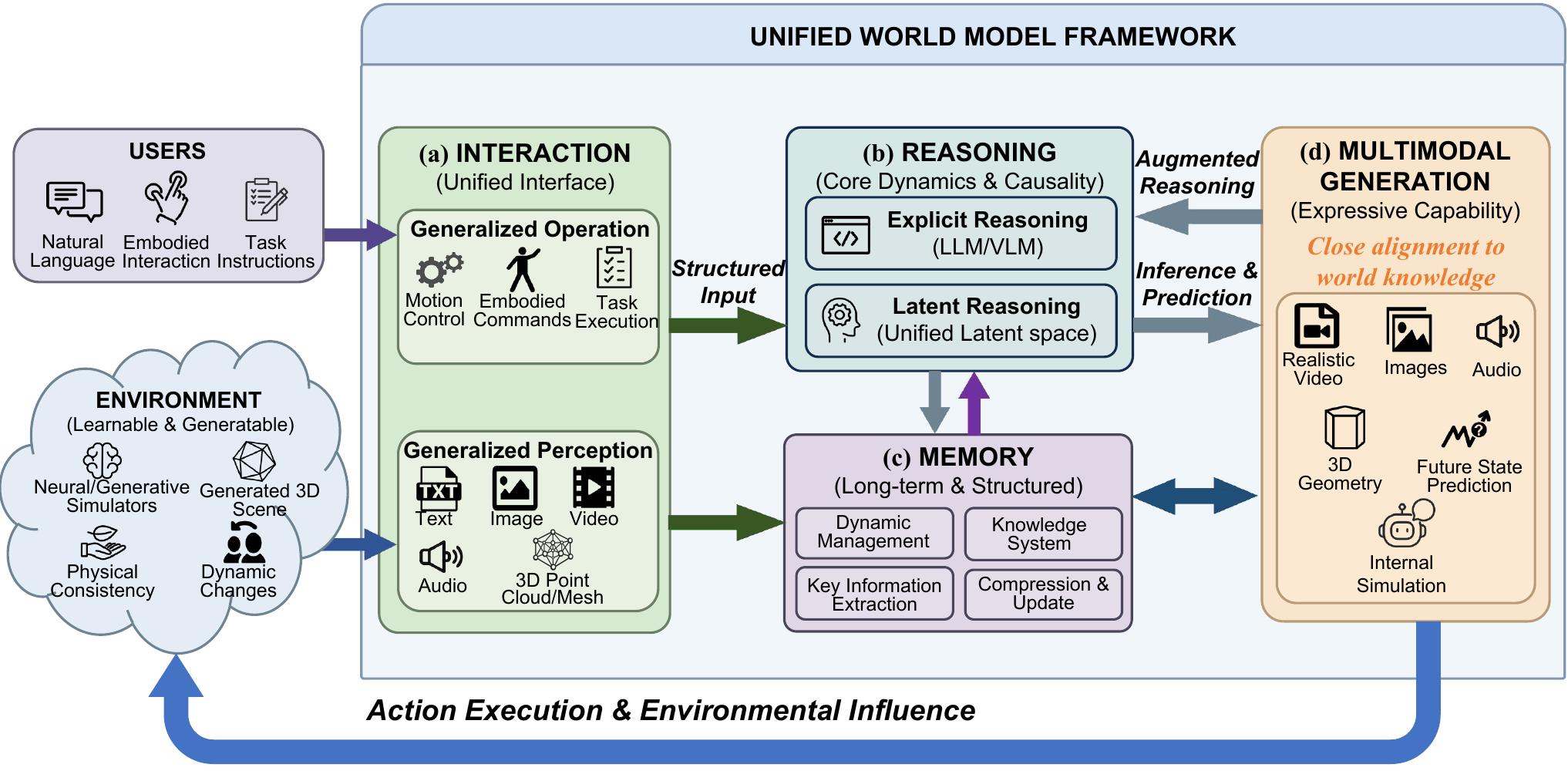}}
    \caption{Illustration of the advocated unified world model framework. Each component served as: (a) Interaction: Enabling the model to handle multi-format inputs from the complex physical world. (b) Reasoning: Conducting logical analysis and inference derived from complex inputs. (c) Memory: Supporting long-term retention and extensive context processing. (d) Multimodal Generation: Empowering the model to generate multimodal outputs, which serve both as environmental feedback and as a catalyst for superior reasoning.}
    \label{fig:framework}
\end{figure*}

\section{Background}

Understanding world knowledge is crucial for enhancing the ability of artificial intelligence systems to handle complex physical environments. Existing research can be broadly categorized into three classes based on the proactivity of their interaction with the environment and their approach to knowledge integration. Although these methods have made progress in their respective fields, they collectively highlight an urgent need for a unified and proactive world modeling framework in current research.

\subsection{Reasoning with World Knowledge}

First, since Large Language Models and Vision-Language Models (LLM/VLM) have demonstrated powerful reasoning and generalization capabilities, some studies have built upon this foundation to further enhance models' reasoning abilities concerning complex physical worlds and challenging logical concepts. This category of work primarily includes: general multimodal reasoning represented by OpenAI O3~\cite{wang2025simple, bai2025multi, liang2025multimodal}, research related to spatial reasoning~\cite{yang2025cambrian, chen2024spatialvlm}, reasoning for challenging competition problems~\cite{chai2025scimaster, qiu2025physics}, and reasoning with multimodal inputs such as audio, 3D, and long videos~\cite{tian2025step, xie2025mini, xie2025audio, liu2025thinksound, shi2025sam, huang2025surprise3d, shi2025mavors, wiedemer2025video, lu2025see4d, chen2025versavid, an2024mc, lin2025perceive, guo2025video}. Meanwhile, with the advancement of reasoning capabilities in large models and agents, some methods~\cite{Park2023Generative, tan2025lumine} have further strengthened interactive capabilities, enabling agents to perform long-term memory and interaction within complex virtual environments. However, despite the already formidable reasoning power of large models, they still face significant challenges in achieving accurate perception of the complex physical world, generating output representations across more modalities, and interacting with the real physical world.

\subsection{World-Driven Content Generation}

In addition to enhancing large language models based on text token prediction by incorporating world knowledge, generative methods in other modalities also actively integrate world knowledge. The earliest attempts to introduce world knowledge into visual generation focused on navigation and abstract reasoning tasks~\cite{bar2025navigation}, where researchers evaluated the generated image sequences or videos to assess the model's accurate cognition of complex spatio-temporal relationships. With the advancement of diffusion models~\cite{liu2022flow, lipman2022flow, flux, peebles2023scalable, li2024zone, shi2025svg, wang2025scone, tong2026cof, an2025unictokens}, the quality of image, video generation and editing~\cite{liu2025javisdit, google2025veo3, openai2025sora2, openai2024sora, alibaba2025wan25, gao2025seedance, zhang2025waver, wan2025wan} has significantly improved. To make the outputs more realistic and reliable, researchers employ techniques such as fine-tuning and reinforcement learning to guide generative models to better adhere to the physical laws of the real world~\cite{li2025hunyuan, tang2025hunyuan, team2025hunyuanworld, zeng2024ipdreamer, zeng2024trans4d, yang2025widerange4d, yang2024semantic, tang2025we, sun2025worldplay, he2025matrix, zhang2025matrix}, aiming to build high-quality ``world simulators''. However, this pixel-estimation-based approach, although richer in information than text token prediction, essentially learns a mapping from the 3D world to 2D rendered results. Even when the generation quality is high, results often violate common sense in details and spatio-temporal logic. Therefore, existing diffusion-based generators do not yet possess a precise understanding of the spatio-temporal relationships in complex physical worlds.

\subsection{Agents in Interactive Environments}

To realize the practical value of world models, research on agent exploration and task execution in autonomous driving, embodied intelligence, and simulated environments is crucial. This line of work aims to integrate world knowledge into the agent's perception-decision loop to achieve more autonomous and physically plausible interactions. For instance, vision-language-action models in robotics~\cite{black2410pi0, bu2025agibot, agarwal2025cosmos, team2025gigabrain}, autonomous driving systems designed for complex decision-making and planning~\cite{hu2023gaia, tu2025role, zeng2025rethinking, russell2025gaia}, and research on training agents to accomplish open-ended tasks in virtual environments~\cite{wang2024omnijarvis, zang2025rlinf} like Minecraft all require models to deeply understand environmental dynamics and perform planning. Although generalist agent frameworks~\cite{black2410pi0, bu2025agibot, team2025gigabrain} have demonstrated the potential to handle multimodal and multitask scenarios, current vision-language-action systems still face limitations in long-term memory, multimodal perception in complex environments, and intricate cross-modal behavioral interactions. This underscores the urgency of a co-designed integration of interaction, perception, reasoning, and memory, which forms a core argument for our advocacy of a unified world model framework.

\section{Unified World Model Framework}

To address the fragmentation in current research and facilitate the development of more robust systems, this section outlines the essential components of a normative world model framework. As illustrated in Fig.~\ref{fig:framework}, the proposed unified framework comprises the following elements.

The original World Models~\cite{ha2018world} primarily consisted of a vision model that receives world inputs, a memory model for dynamic prediction and processing, and a controller that governs the model's outputs. This established an effective foundational architecture for the world model framework. However, with advancements in fields such as LLMs/VLMs, diffusion models, and VLAs, this basic framework requires further expansion and refinement.

\paragraph{Interaction.} The fundamental value of a world model lies in its ability to engage in bidirectional, multimodal interactions with complex environments and users. Consequently, its interaction module should evolve beyond the early framework's ``vision model'', and advance into a unified perceptual and operational interface. As shown in Fig.~\ref{fig:framework}(a), This interface requires two core capabilities: first, generalized perception, enabling the understanding and processing of multimodal inputs such as text, images, video, audio, 3D point clouds, and meshes to form a unified representation of the world state; second, generalized operation, allowing the parsing and execution of diverse task instructions. These instructions include not only natural language or embodied interaction commands from users, such as movement, rotation, or dragging, but also low-level motion control signals for agents like robots or vehicles. To achieve efficient and reliable closed-loop interaction, the world model's interaction module must unify the scheduling, encoding, and organization of these heterogeneous perceptual data and operational signals, providing structured input for subsequent reasoning, memory, and generation.

\paragraph{Reasoning.} To navigate the complex and dynamic nature of the real world, a world model necessitates a core component dedicated to reasoning about intricate dynamics and causality. Currently, LLMs/VLMs integrated with Explicit Reasoning have demonstrated remarkable analytical capabilities. A mainstream and effective strategy is to employ them within a world model, as illustrated in Fig.~\ref{fig:framework}(b). Explicit Reasoning transforms multimodal observations and interactive information into textual descriptions or reasoning chains, leveraging the powerful symbolic reasoning and planning abilities of LLMs to infer physical laws, predict future states, or formulate high-level strategies. This text-mediated reasoning offers high transparency and is relatively easy to align and verify with human intuition. For scenarios requiring the handling of sub-symbolic and continuous physical details, Explicit Reasoning may lead to information loss, making the introduction of Latent Reasoning more appropriate. This approach would enable reasoning directly within a unified latent space, jointly leveraging encoded multimodal information from vision, language, action, etc. Regardless of the algorithmic approach, the reasoning module of a world model should fundamentally possess the capability to perform rational inference on inputs, generating more structured and coherent content.

\paragraph{Memory.} To maintain coherence and consistency in complex, continuous physical tasks, a world model must possess robust long-term memory capabilities. Memory mechanisms have evolved from implicit state storage based on recurrent networks like LSTMs~\cite{hochreiter1997long,beck2024xlstm} to explicit large-scale memory utilizing the Transformer architecture with long-context windows~\cite{beltagy2020longformer,dao2022flashattention,dao2023flashattention,ji2025memflow}. As illustrated in Fig.~\ref{fig:framework}(c), Faced with multimodal, high-concurrency interaction streams in an open world, the memory module of a world model must transcend simple sequential storage and achieve structured and dynamic management of information. This requires the system to effectively categorize, associate, and fuse experiential data from different modalities and sources, thereby constructing a unified and queryable internal knowledge system. Simultaneously, constrained by computational resources, the memory system must possess the capability for key information extraction and compression~\cite{yang2025cambrian}, actively filtering and retaining states and events core to the task. Furthermore, memory should be a dynamically evolving process, as interactions progress, the system must continuously merge, update, and purge redundant stored content to ensure its timeliness and conciseness.

\begin{figure*}[h]
    \centering
    \centerline{\includegraphics[width=\linewidth]{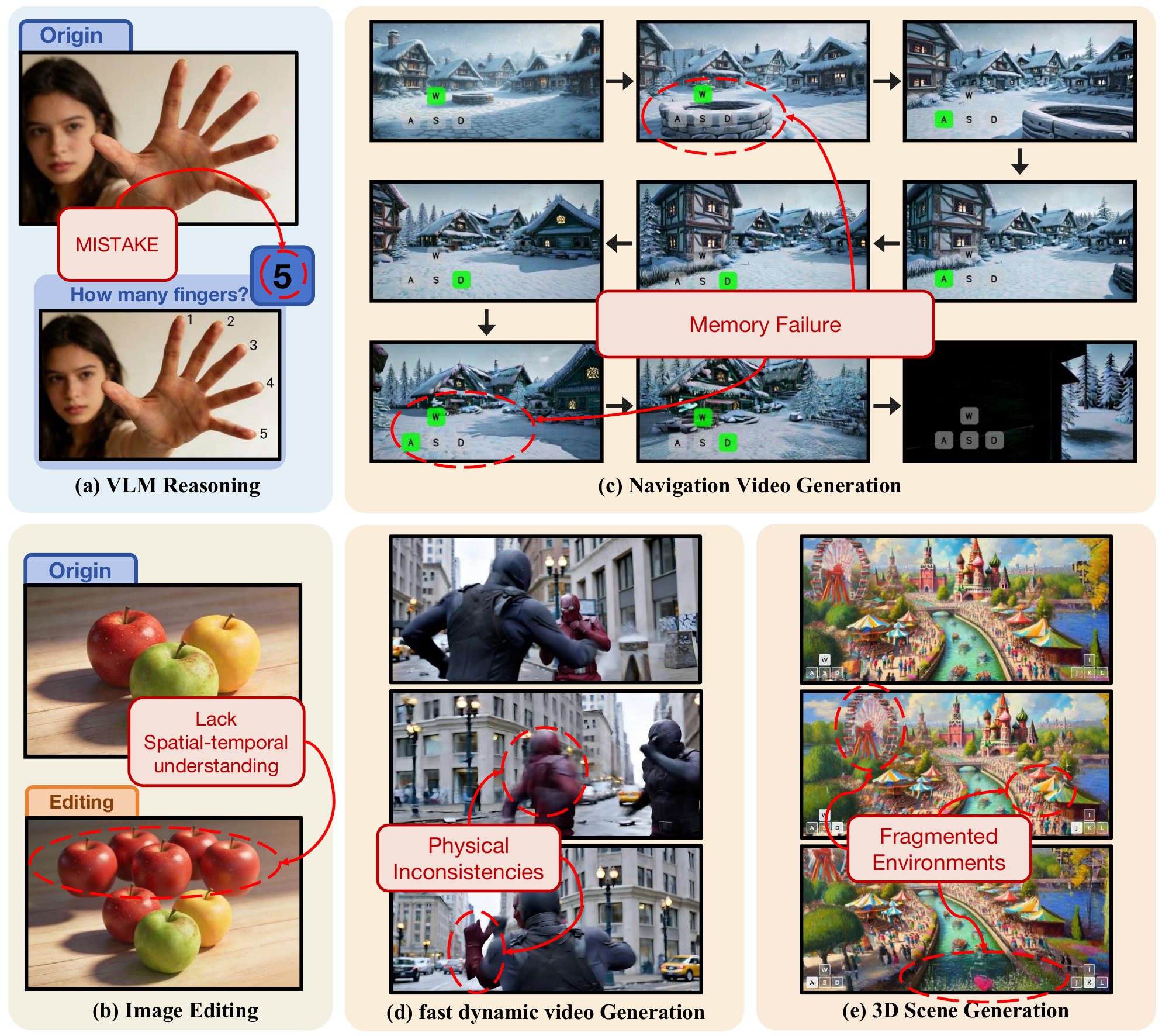}}
    \caption{Failure cases of various task-specific methods infused with world knowledge.}
    \label{fig:challenge}
\end{figure*}

\paragraph{Environment.} The training and validation of a world model are inseparable from an interactive and controllable environmental carrier. We posit that the environment should encompass both the complex physical world and simulation environments, while simultaneously serving as an integral part of the world model that receives and updates based on outputs from other components, as illustrated on the left side of Fig.~\ref{fig:framework}. Traditionally, high-cost interaction with physical hardware~\cite{black2410pi0,hu2023gaia} is the ultimate goal for world models to engage with the real world, however, acquiring training data at scale remains a significant challenge. Simulation environments~\cite{kolve2017ai2,li2022metadrive,zhang2025autoenv} have served as the cornerstone of early-stage research by providing controllable, safe, and efficient physical or rule-based simulation. However, most simulation environments rely on manually modeled limited scenes and rigid-body dynamics, creating a ``sim-to-real'' gap in terms of authenticity and diversity. We advocate that the environmental architecture for world models should possess generative and extensible capabilities. Specifically, techniques such as 3D generation methods~\cite{li2025flashworld,yu2025wonderworld} and procedural content generation should be leveraged to dynamically synthesize near-infinite, high-fidelity virtual scenes. Such a generative environment should function not merely as a scene ``renderer'' but as a physically consistent simulator capable of responding to complex interactions and producing dynamic changes conforming to real-world laws. This would enable world models to be trained on an extremely rich and realistic distribution of environments, enhancing their generalization and adaptation capabilities for open, unknown real-world scenarios.

\paragraph{Multimodal Generation.} While accepting complex inputs and performing reasoning, the world model must also possess multimodal generation capabilities to provide comprehensive feedback on complex environmental changes. This is a crucial capability and a vital means of verifying the accuracy of its world understanding and achieving intuitive alignment with humans. As shown in Fig.~\ref{fig:framework}(d), a complete multimodal generation capability for a world model should extend beyond generating textual reports, it must be able to generate realistic video, images, audio, and even 3D geometry based on its internal states and future predictions. For instance, in an embodied navigation task, after receiving instructions and initial observations, the model should be able to synthesize a 3D scene from the agent's perspective based on a 3D representation, such as point clouds. This constitutes an internal simulation of its own navigation strategy and scene comprehension. Multimodal generation should not be an isolated output module but should form a closed loop with the reasoning and memory modules. Generated scenes can provide model-based foresight for planning, and generated data can be utilized for self-augmentation, continuously refining and enriching the model's world knowledge.

\section{Limitations of Existing Models Incorporating World Knowledge}

This section analyzes the limitations inherent in current approaches across different domains, substantiating the need for the integrated framework proposed above.

For the most widely applied Large Language Models (LLMs) and Vision-Language Models (VLMs), although these models appear to possess extensive world knowledge, they fundamentally rely on statistical fitting of large-scale training data. This limitation becomes evident in complex academic reasoning, such as failing to accurately recognize chemical formulas in Chemistry Olympiad problems, and in counter-intuitive multimodal recognition. As shown in Fig.\ref{fig:challenge}(a), when an unnatural image depicting six fingers is input into a large model, it may still assert that there are only five fingers in the picture. This indicates that large models are heavily influenced by large-scale training data and struggle to discern irregular or unnatural scenarios. These shortcomings of LLMs and VLMs suggest a lack of effective perception of real-world complexity and a genuine understanding of physical laws. We argue that accurately representing multimodal inputs within a spatial and physical framework would significantly enhance the models' comprehension of the world.

Regarding image generation and editing, early methods like AnyEdit~\cite{yu2025anyedit} and EditWorld~\cite{zeng2025editworld} primarily focused on curating task-specific datasets enriched with world knowledge to improve editing performance. However, training diffusion models directly on such data often fails to handle complex, logic-heavy instructions. Conversely, frameworks that integrate VLMs with diffusion processes have demonstrated superior representational capabilities compared to data-centric methods. This reinforces our argument that architectural advancement is more promising than mere data injection. Current editing methods still lack effective interaction with the physical world and spatio-temporal understanding. As shown in Fig.\ref{fig:challenge}(b), although the model successfully completes the editing task, the results do not conform to real-world lighting and shadow patterns. This indicates that possessing only logical reasoning and image generation capabilities is insufficient to produce images that align with real-world dynamics. Effectively capturing the complex, rule-based changes of the physical world remains crucial for models. In summary, developing a comprehensive world model framework represents a viable strategy for advancing image generation and editing.

In video generation, navigation video synthesis is frequently cited as a key capability of world models~\cite{li2025hunyuan, zhang2025matrix, zhu2025astra, bahmani2025lyra, ding2025kling, wan2025wan}. Although these models aim to function as world simulators, they often struggle with long-term memory management. As illustrated in Fig.\ref{fig:challenge}(c), when moving left for a certain distance and then returning to the right, the objects originally present in the scene noticeably disappear, which clearly violates physical laws, this indicates that these models are merely focused on next-frame prediction in video generation, lacking effective long-term memory and real-world understanding capabilities. Furthermore, we demonstrate the performance of existing state-of-the-art generative models in Fig.~\ref{fig:challenge}(d); despite their high visual quality, their outputs fail to align with real-world principles when synthesizing complex, high-speed dynamic videos. These approaches continue to fit pixel-level patterns rather than internalizing the underlying laws of the world, leading to physical inconsistencies over time.

Current 3D generation methods suffer from inadequate dynamics and scalability. The resulting 3D outputs often achieve only ``visual plausibility'' without possessing genuine physical significance or interactive properties. Moreover, constrained by computational limits, directly generated 3D spaces are frequently limited in scale, leading to fragmented environments. As illustrated in Fig.\ref{fig:challenge}(e), although the overall quality of the 3D scene generated by existing methods appears high, details exhibit noticeable fragmentation and distortion due to the limited representational capacity of 3D point clouds~\cite{chen2025sam, huang2025midi}. This further demonstrates that current 3D generation approaches remain at the level of visual alignment and struggle to handle complex 3D spaces. Merely improving memory strategies is still insufficient for capturing the laws of the real physical world. Therefore, by holistically enhancing the memory, multimodal generation, and reasoning components within a world model, 3D synthesis could transcend current spatial limitations and better align with the evolutionary principles of the complex world.

\begin{figure}[t]
    \centering
    \centerline{\includegraphics[width=\columnwidth]{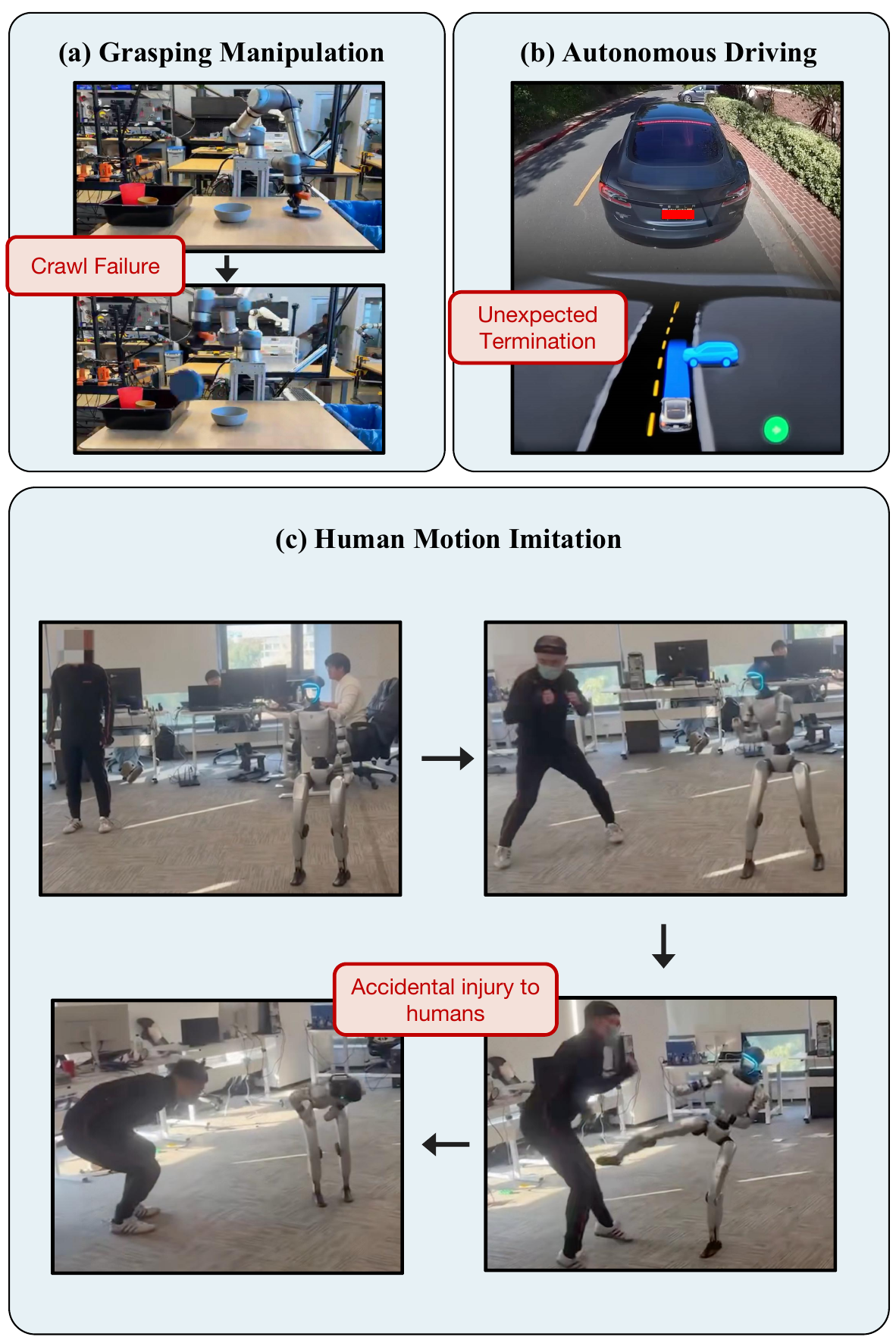}}
    \caption{Illustration of the limitations of existing embodied AI and autonomous driving systems. \textit{Images sourced from internet search}.}
    \label{fig:embody_limit}
    \vspace{-4mm}
\end{figure}

Finally, for autonomous driving and embodied AI, while the integration of world knowledge has yielded performance gains, these methods remain confined to narrow, task-specific domains. They often lack a deep understanding of complex, long-horizon multimodal contexts. As shown in Fig.~\ref{fig:embody_limit}(a), current mainstream embodied AI research typically combines robotic arms with recognition and reasoning models to accomplish simple planning tasks. However, these tasks remain relatively basic and fail to evaluate the model's capabilities in real-world complex scenarios. Meanwhile, although some efforts have deployed autonomous driving and embodied intelligence in practical applications, these achievements exhibit notable instabilities. Fig.~\ref{fig:embody_limit}(b) illustrates cases where autonomous vehicles fail to handle relatively straightforward road conditions, highlighting the considerable gap that remains before such systems can adeptly navigate complex real-world environments. Similarly, Fig.~\ref{fig:embody_limit}(c) presents a robot that, despite being capable of imitating human movements with reasonable fidelity, inadvertently harms a human due to its inability to deviate from pre-programmed actions. These examples collectively demonstrate that merely integrating existing models with embodied systems only enables the execution of basic, pre-defined tasks. We posit that autonomous driving and embodied agents should serve as the ``carriers'' for a world model to explore the environment, while high-quality control should be an emergent capability of the model itself. Simply coupling large models with physical hardware~\cite{li2025large} to improve task success rates deviates from the original objective of world models: to create agents capable of active exploration, discovery, and response to complex environments.

\section{Discussion: Standardization and Feasibility}

The proposal for a unified world model framework invites discussion regarding feasibility and the trade-offs with task-specific optimization.

\paragraph{Efficiency vs. Generalization.} A prevailing perspective is that fine-tuning specialized models for specific tasks (e.g., robotic grasping) yields optimal performance with clear engineering paths. Indeed, unified frameworks may incur higher training costs and complexity compared to highly optimized, task-specific systems. However, this view focuses on static performance metrics. From the perspective of dynamic interaction in open-ended environments, task-specific models often hit a performance ceiling defined by their training data. A unified framework offers the structural foundation for knowledge transfer between tasks and lifelong learning, which are essential for general world understanding.

\paragraph{Diversity vs. Integration.} Another consideration is whether a unified framework might stifle technological diversity. It can be argued that sub-problems like perception and reasoning are distinct and require specialized architectures. However, the ``unification'' proposed here does not imply a rigid, monolithic network. Instead, it advocates modular functional specifications and standardized interfaces. By defining how core components (interaction, memory, reasoning) collaborate, a standardized framework can facilitate the integration and benchmarking of diverse research efforts. This approach aims to redirect focus from redundant low-level developments to high-level system optimization, potentially accelerating the field's overall advancement.

\section{Future Work}

After analyzing the current research field of World Models and proposed a unified normative framework, this section explores several critical directions that are essential for future breakthroughs in the field.

\paragraph{Physically-Grounded Spatiotemporal Representation.}
Precise perception and reconstruction of temporal-spatial environment serve as the cornerstone for reasoning and generation within World Models. However, existing 3D and 4D representation techniques still face formidable challenges. While methods such as 3D Mesh, NeRF~\cite{mildenhall2021nerf}, 3D Gaussian Splatting~\cite{kerbl20233dgs}, and 4D representation models~\cite{wu20244dgs, yang2023realtime4dgs} have made significant strides in fitting visual appearances, enabling the synthesis of photorealistic objects or scenes, they remain essentially optical representations. They lack an intrinsic expression of real-world physical properties, such as mass, friction, elasticity, and collision volume. Furthermore, current representations struggle to support free exploration and interaction under low computational overhead. For instance, 3DGS often relies on massive point clouds to force-fit visual effects, such unstructured, discrete representations are difficult to map onto physically consistent entities, leading to logical fallacies when the model handles object deformation, fluid dynamics, or complex contacts. Consequently, future research must transcend mere appearance reconstruction and pivot toward physically-grounded representation. We need to explore novel data structures or neural implicit representations that embed physical attributes while maintaining high-fidelity visuals, significantly reducing the computational cost of rendering and interaction. This will provide World Models with a spatiotemporal representation that is both freely explorable and strictly compliant with physical laws.

\paragraph{Embodied Interaction and Control.}
Embodied AI serves as the ideal vehicle for World Models to explore and validate their understanding of the real world. The current bottleneck lies in the difficulty of directly transferring policies generated by World Models to physical robots, a limitation rooted in the operational flexibility, sensing precision, and physical plausibility of current embodied systems. Future development should focus on enhancing the control capabilities of World Models within complex, dynamic environments. First, models must adapt to robot morphologies with higher Degrees of Freedom (DoF), extending from simple grasping tasks to fine-grained dexterous manipulation. Second, the Sim-to-Real Gap must be bridged, enabling World Models generate action sequences that respect hardware constraints, such as torque limits and joint singularities, thereby allowing embodied agents to effectively navigate diverse real-world scenarios. Furthermore, World Models should be endowed with long-horizon planning capabilities, allowing them to comprehend the causal logic of tasks and command embodied agents to complete multi-stage, complex missions in unstructured environments. Ultimately, after understanding the world, the model should be able to perform sophisticated tasks in the real-world through physical robotic platforms.

\paragraph{Autonomous Reflection and Modular Continuous Evolution.}
Beyond enhancing the capacity for external exploration, improvements to the the World Model itself are equally crucial. Current systems rely heavily on offline large-scale training and lack mechanisms for active error correction or self-updating post-deployment. Future research should strive to empower World Models with metacognition and self-reflection. Specifically, models should possess the ability for uncertainty estimation regarding their own predictions. When a significant discrepancy arises between a prediction and actual observation, the model should autonomously trigger a reflection mechanism to identify knowledge gaps. It should then spontaneously perform targeted fine-tuning by collecting specific data or replaying high-value samples, rather than passively awaiting a full retraining cycle. Although current Reinforcement Learning methods assist in proactive thinking, they remain tethered to human-defined reward functions. Thus, achieving autonomous exploration within the World Model is essential. Simultaneously, to meet evolving task requirements, World Models must feature efficient and flexible modular iteration. Modules for perception, memory, reasoning, and planning should support independent fine-tuning and upgrades. This design allows researchers to iteratively improve specific weaknesses (e.g., upgrading the physical reasoning module without impairing the World Models' other capabilities), thereby achieving lifelong learning and agile evolution of the entire system.

\section{Conclusion}
In this paper, we have analyzed the current state of world model research, noting a prevalence of task-specific integrations. While valuable, these approaches often lack the systemic coherence necessary for general world understanding. We proposed a Unified World Model Framework that integrates interaction, perception, reasoning, memory, and generation into a normative design. By discussing the limitations of existing methods and the trade-offs of standardization, we highlight the potential of this framework to foster more robust and principled research. We hope this work serves as a guideline for future endeavors in physically-grounded representation, embodied control, and autonomous evolution, ultimately advancing agents capable of active and intelligent interaction with the complex world.


\section*{Impact Statement}

This paper advocates for a unified framework in world model research to enhance reproducibility and robustness. While the proposed theoretical framework poses no direct societal harm, we acknowledge the risks associated with advanced world models, including the generation of misleading content and safety issues in embodied agents. We emphasize the importance of embedding ethical considerations and safety-by-design principles in the development of these systems to ensure beneficial and secure outcomes.

\bibliography{main}
\bibliographystyle{icml2026}

\end{document}